\documentclass[journal]{IEEEtran}
\usepackage{bm}
\usepackage{subfigure}
\usepackage{amsmath,amssymb,amsthm}
\usepackage{multirow,array}
\usepackage{color}
\usepackage{algorithmic,algorithm}
\usepackage{lipsum}
\usepackage{booktabs}
\usepackage{graphicx}

\newcommand{\paraheading}[1]{\vspace{1em} \noindent \textbf{#1} \hspace{0.2em}}

\usepackage{bm}
 \usepackage{cite}
\usepackage{subfigure}
\usepackage{amsmath,amssymb,amsthm}
\usepackage{multirow,array}
\usepackage{color}
\DeclareMathOperator{\Tr}{Tr}
\DeclareMathOperator{\VEC}{VEC}
\newtheorem{theorem}{Theorem}

\newtheorem{proposition}{Proposition}

\begin{document}
\title{\textcolor[rgb]{0.00,0.00,1.00}{A Multilayer Framework for Online Metric Learning}}

\author{Wenbin~Li,
        Yanfang~Liu,
        Jing~Huo,
        Yinghuan~Shi,
        Yang~Gao,~\IEEEmembership{Senior Member,~IEEE},
        Lei~Wang,~\IEEEmembership{Senior Member,~IEEE},
        and~Jiebo~Luo,~\IEEEmembership{Fellow,~IEEE}
\thanks{W. Li, J. Huo, Y. Shi and Y. Gao are with the State Key Laboratory for Novel Software Technology, Nanjing University, Nanjing 210023, China. (e-mail: liwenbin@nju.edu.cn; huojing@nju.edu.cn; 
syh@nju.edu.cn; gaoy@nju.edu.cn).}
\thanks{Y. Liu is  with the State Key Laboratory for Novel Software Technology, Nanjing University, Nanjing 210023, China, and the College of Mathematics and Information Engineering, Longyan University, 364012, China. (e-mail: liuyanfang003@163.com).}
\thanks{L. Wang is with the School of Computing and Information Technology, University of Wollongong, Australia (e-mail: leiw@uow.edu.au).}
\thanks{J. Luo is with the Department of Computer Science, University of Rochester, Rochester, NY 14611, USA (e-mail: jluo@cs.rochester.edu).}
\thanks{Wenbin Li and Yanfang Liu contributed equally as co-first authors.}
\thanks{Corresponding author: Yang Gao.}}


\maketitle

\begin{abstract}
Online metric learning has been widely applied in classification and retrieval. It can automatically learn a suitable metric from data by restricting similar instances to be separated from dissimilar instances with a given margin. However, the existing online metric learning algorithms have limited performance in real-world classifications, especially when data distributions are complex.
To this end, this paper proposes \textcolor[rgb]{0.00,0.00,1.00}{a multilayer framework for online metric learning} to capture the nonlinear similarities among instances.
Different from the traditional online metric learning, which can only learn one metric space, \emph{multi-layer online metric learning (MLOML)} takes an online metric learning algorithm as a metric layer and learns multiple hierarchical metric spaces, where each metric layer follows a nonlinear layers for the complicated data distribution.
Moreover, the forward propagation (FP) strategy and backward propagation (BP) strategy are employed to train the metric layers.
To build a metric layer of the proposed MLOML, a \textit{Mahalanobis-based Online Metric Learning (MOML)} algorithm is presented based on the passive-aggressive strategy and one-pass triplet construction strategy.
Furthermore, in a progressively and nonlinearly learning way, MLOML has a stronger learning ability than traditional online metric learning in the case of limited available training data. 
To make the learning process more explainable and theoretically guaranteed, theoretical analysis is provided. 
The proposed MLOML enjoys several nice properties, indeed learns a metric progressively, and performs better on the benchmark datasets.
Extensive experiments with different settings have been conducted to verify these properties of the proposed MLOML. 
\end{abstract}


%
\IEEEpeerreviewmaketitle

\section{Introduction}

Learning a meaningful and quality metric on the original instances is crucial to many classification and retrieval applications.
In recent decades, many metric learning methods based on Mahalanobis distance function and bilinear similarity function have been proposed.
Mahalanobis distance-based methods~\cite{xing2002distance,weinberger2005distance,nguyen2019kernel,davis2007information,xiang2008learning,ying2018manifold,ye2019fast,liu2020logdet} refer to learning a real-valued distance matrix with a symmetric positive semi-definite (PSD) constraint.
Bilinear similarity-based methods~\cite{chechik2010large,gao2014soml,xia2014online} aim to learn a form of bilinear similarity matrix without the PSD constraint.
Moreover, there are two kinds of constraints, i.e., pairwise and triplet constraints, that have been widely used in these metric learning methods. 
A pairwise constraint consists of two similar or dissimilar instances, while a triplet constraint is of the form $\langle\bm{x},\bm{x}^+,\bm{x}^-\rangle$, where instance $\bm{x}$ is similar to instance $\bm{x}^+$, but is dissimilar to instance $\bm{x}^-$.

In many real-world applications, a lot of data is streaming data which is continuously produced in time, such as wind power data~\cite{kavousi2015new}, credit data~\cite{duhon2001line}, and ADs click data~\cite{he2014practical}. 
Online learning algorithm investigates how to learn in a streaming setting~\cite{crammer2006online,yu2020online,yu2022online}.
Therefore, metric learning algorithms should be able to learn metric in an online manner, \textit{i.e., online metric learning (OML)}.
In fact, multiple OML algorithms have been proposed~\cite{shalev2004online,jain2009online,jin2009regularized,chechik2010large,li2017opml,gao2019towards}. 
However, the existing OML algorithms mainly pay attention to rapid constraints construction~\cite{chechik2010large,li2017opml} or low update complexity~\cite{jain2009online,jin2009regularized,li2017opml}, while rarely consider the learning ability in the case where all the labeled streaming data cannot be observed. In addition, most of these OML algorithms only learn one linear metric space which cannot learn well refined metrics for a complicated nonlinear data distribution.

\begin{figure}[!tbp]
  \centering
  \includegraphics[width=0.47\textwidth]{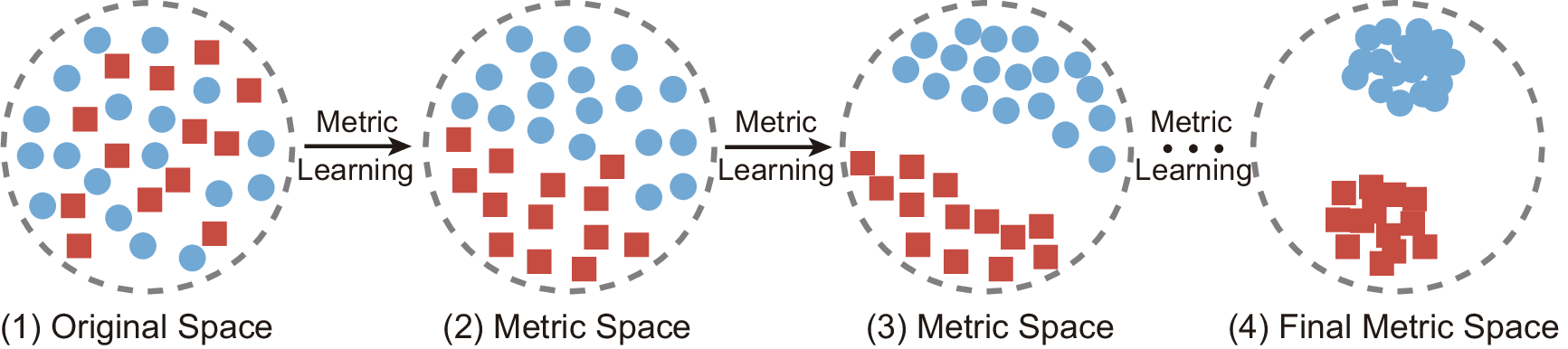}
  \caption{An illustration of online hierarchical metric learning by learning new metric spaces progressively.}
  \label{fig-ODML-idea}
\end{figure} 

To tackle the above limitations of the existing OML algorithms, we propose a \emph{Multi-Layer Online Metric Learning (MLOML)} framework, which is nonlinear.
In this framework, we attempt to design a \textit{metric-algorithm-based layer}, which is stacked by several OML algorithms along with the corresponding nonlinear layers (\emph{e.g.,} ReLU, Sigmoid, tanh). 
In this way, our proposed MLOML is able to learn a progressively refined metric space by learning another new metric in the former learnt feature space (see Fig.~\ref{fig-ODML-idea}). Specifically, in MLOML, one OML algorithm is taken as a metric layer, followed by a nonlinear layer (\emph{i.e.,} ReLU, Sigmoid, or tanh). 
These two layers are repeatedly stacked multiple times. 
It is worth noting that each metric layer in MLOML is a relatively independent OML algorithm, as a result, the parameters of each metric layer can be innovatively updated according to its own local loss during forward propagation (FP). 
It means that it is possible to train such a metric-algorithm-based layer by only using the FP strategy. 
The advantages of this FP updating are: (1) the parameter updating is immediate, unlike the delayed updating of the commonly used backward propagation (BP); (2) when additional BP is adopted, FP updating can vastly accelerate the convergence. Note that the second advantage has a similar effect of layer-wise unsupervised pre-training~\cite{hinton2006reducing,bengio2007greedy,erhan2010does}. 
However, there are fundamental differences. 
The existing layer-wise training is unsupervised and only acts as a pre-training operation (or a regularizer~\cite{erhan2010does}), which is not end-to-end. 
In contrast, the FP updating in the proposed MLOML is supervised and serves the primary training mode rather than a pre-training role (elaborated in Section~\ref{FP-BP-FBP}), which is end-to-end. In fact, these two updating strategies (\emph{i.e.,} FP and BP) can be combined to train this metric-algorithm-based layer. 
Ideally, FP updating can explore new feature spaces sequentially, while BP updating can amend the exploration in further.

Furthermore, to achieve a low computational cost when performing MLOML, a new general \textit{Mahalanobis-based Online Metric Learning (MOML)} algorithm is proposed as the metric layer of MLOML.
Since all the labeled streaming data cannot be observed in online manner, MOML uses the one-pass triplet construction~\cite{li2017opml} Instead of triple constraints obtained in advance.
Simultaneously, MOML has a convex objective function inspired by passive-aggressive learning and enjoys a closed-form solution at each step. 
We also derive a theoretical regret bound for MOML to prove its convergence. Through stacking MOML hierarchically, the ability of learning feature representation progressively can be explainable and guaranteed. 

Our main contributions can be summarized as follows: 
\begin{itemize}
    \item A \emph{Multi-Layer Online Metric Learning (MLOML)} framework is developed for streaming data through forward propagation (FP) strategy or backward propagation (BP) strategy, such that a metric space is learned progressively and deeply, \emph{i.e.,} exploring and learning a new metric in a nonlinear transformation space sequentially. 
    \item Taking \emph{Mahalanobis-based Online Metric Learning (MOML)} as a metric layer, MLOML has theoretical guarantees so that the classification performance will be improved or at least well maintained as the depth of the layers increases. 
    \item MLOML is simple yet effective, as verified by extensive experiments.
\end{itemize}

\section{Related work}
Online metric learning enjoys several practical and theoretical advantages, making it widely studied and applied in data mining tasks, which can be roughly divided into two categories: Mahalanobis distance-based and bilinear similarity-based methods.
In bilinear similarity-based method, Online Algorithm for Scalable Image Similarity (OASIS)~\cite{chechik2010large} is proposed based on PA algorithm, aiming to learn a similarity metric without PSD constraint. 
Following a similar setting as OASIS, Sparse Online Metric Learning (SOML)~\cite{gao2014soml} learns a diagonal matrix instead of a full matrix to deal with the high-dimensional data. 
Online Multiple Kernel Similarity (OMKS)~\cite{xia2014online} has been proposed to handle the multi-modal data.
Through adopting an off-diagonal $\ell_1$ norm to the similarity matrix, Sparse Online Relative Similarity (SORS)~\cite{yao2015sparse} can obtain a sparse result.
Online Similarity Learning via Low Rank and Group Sparsity (OSLLR-GS)~\cite{cong2018online} is designed to address the over-fitting problem for big data by detecting the feature redundant in the metric matrix and constraining the remaining matrix to a low rank space.

\begin{figure*}[!t]
  \centering
  \includegraphics[width=0.65\textwidth]{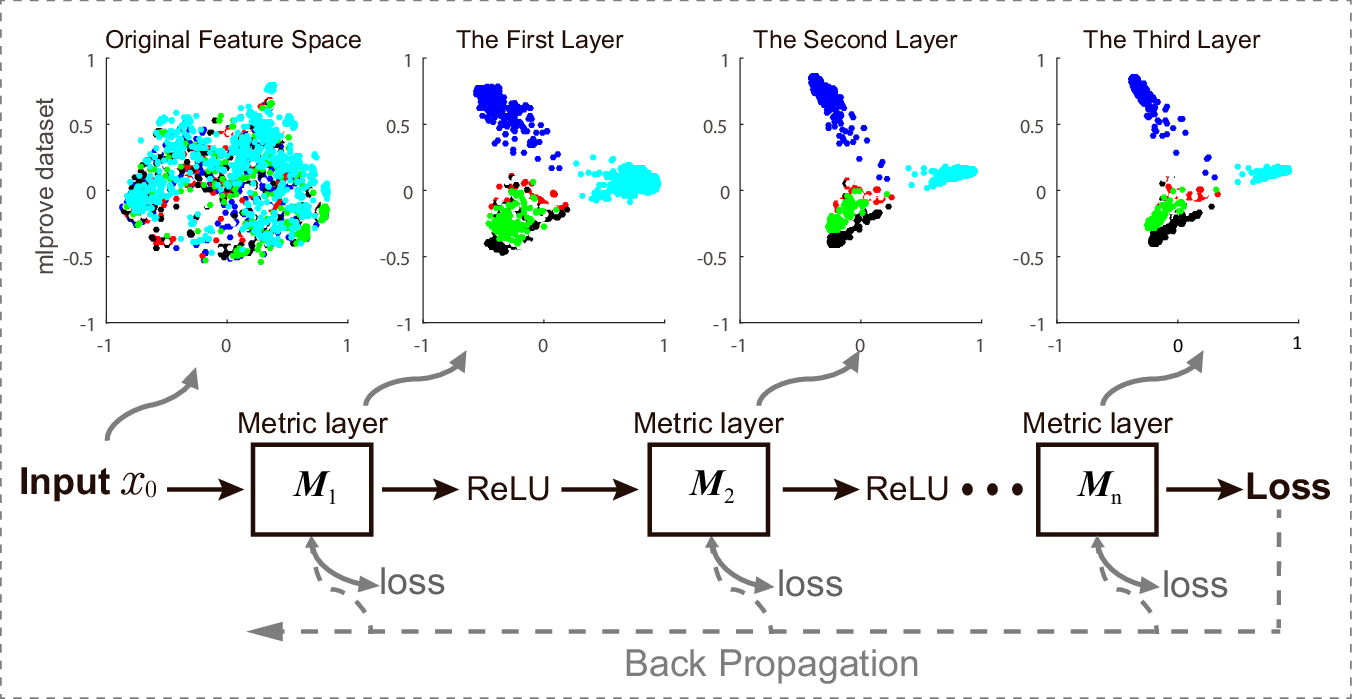}\\
  \caption{Framework of the proposed multi-layer online metric learning (MLOML), where each metric layer $\bm{M}_n$ is an online metric learning algorithm.
  Here we take three metric layers and two ReLU layers as an example, \emph{i.e.,} $n = 3$.}
  \label{fig-flowchart}
\end{figure*}

In the second kind of Mahalanobis distance-based methods, Pseudo-Metric Online Learning Algorithm (POLA)~\cite{shalev2004online} introduces the successive projection operation to update a pseudo-metric and map it onto a positive semi-definite cone. 
As an extended version of Information Theoretic Metric Learning-Online (ITML-Online)~\cite{davis2007information}, LogDet Exact Gradient Online (LEGO)~\cite{jain2009online} updates a manalanobis metric based on LogDet regularization and gradient descent. 
Regularized Distance Metric Learning (RDML)~\cite{jin2009regularized} with appropriate constraints has a provable regret bound.
Mirror Descent for Metric Learning (MDML)~\cite{kunapuli2012mirror} is an unified approach which updates a manalanobis metric by composite objective mirror descent.
Bellet and Habrard~\cite{bellet2015robustness} utilize an adaptation of the notion of algorithmic robustness~\cite{xu2012robustness} to derive generalization bounds for metric learning.
Low-Rank Similarity Metric Learning (LRSM)~\cite{liu2015low} uses SVD-based projection to solve the challenging high-dimensional learning task, and then employs Alternating Direction Method of Multipliers (ADMM)~\cite{boyd2011distributed} to optimize the model.
Based on Passive-Aggressive (PA) algorithm~\cite{crammer2006online}, Scalable Large Margin Online Metric Learning (SLMOML)~\cite{zhong2018slmoml} adopts the LogDet divergence to maintain the closeness between two successively learned Mahalanobis matrices, and utilizes the hinge loss to enforce a large margin between relatively dissimilar samples.
Fast Low-Rank Metric Learning (FLRML)~\cite{liu2019fast} is an unconstrained optimization on the Stiefel manifold to handle datasets with both high dimensions and large numbers of instances.
Large-Margin Distance Metric Learning (LMDML)~\cite{nguyen2020scalable} employs the principle of margin maximization and stochastic gradient descent method to learn the distance metric with PSD constraint.
These methods almost assume that the pairwise or triplet constraints can be obtained in advance except RDML, which exactly receives two adjacent samples as a pairwise constraints at each time.
In view of adapting the pairwise and triplet constraints to streaming data, Li \emph{et al.}~\cite{li2017opml} present a one-pass triplet construction strategy and design OPML and COPML algorithms with low time complexity. 
By incorporating with a smoothed Wasserstein metric distance, Evolving Metric Learning (EML)~\cite{dong2021evolving} can handle the instance and feature evolutions simultaneously.

However, the above mentioned metric algorithms only learn one linear metric space which cannot learn well refined metrics for a complicated nonlinear data distribution.
In order to solve this issue, we propose MLOML which is developed based on a newly designed Mahalanobis-based online metric learning (MOML).
Compared with the above OML algorithms, MLOML has the following advantages: (1) MLOML is hierarchical and can learn feature representation progressively (\emph{i.e.,} better and better) through FP and BP strategies; (2) MLOML not only has theoretical guarantees by stacking MOML algorithm as its metric layer, but also is nonlinear by employing nonlinear functions; (3) MLOML enjoys a stronger learning ability than traditional OML algorithms with the same amount of data.

\section{Our Framework}
Our goal is to design a novel multi-layer online metric learning framework (MLOML) for streaming data, which is stacked by metric-algorithm-based layers along with the corresponding nonlinear layers (\emph{e.g.,} ReLU, Sigmoid, tanh).
The framework is illustrated in Fig.~\ref{fig-flowchart}.

\subsection{Multi-Layer Online Metric Learning}
In this section, we propose and explain our MLOML in detail. 
MLOML is made up of multiple metric layers and nonlinear layers, in which one metric layer is an OML algorithm and one nonlinear layer is ReLU, Sigmoid or tanh. 
To ensure the progressively learning ability of MLOML, we should guarantee the convexity of each metric layer, which can easily guarantee the convergence of each layer. 
Therefore, a new \textit{Mahalanobis-based OML algorithm (MOML)} is designed specifically. 
MOML has a convex objective function and enjoys a closed-form solution. 
Moreover, a tight regret bound of MOML is also proved (see Theorem~\ref{theorem2}).

Specifically, MOML is built on triplet-based constraints $\langle\bm{x},\bm{x}^+,\bm{x}^-\rangle$, where instance $\bm{x}$ is similar to instance $\bm{x}^+$, but is dissimilar to instance $\bm{x}^-$, and these triplets can encode the proximity comparison information.
Thus, MLOML is also learnt from triplet constraints. 
For computational efficiency, a one-pass triplet construction strategy presented by OPML~\cite{li2017opml} is also employed to construct triplets rapidly which can solve the inability to observe all the streaming data and its labels.
In brief, for each new sample, two latest samples from both the same and different classes in the past samples are selected. 
By using this strategy, triplets can be constructed in an online manner. 
There are two types of layers in MLOML, that are OML layer and non-linear layer, where MLOML-r, MLOML-s, MLOML-t correspond to MLOML with the ReLU, Sigmoid, tanh layers respectively. 
If we design a three-layer MLOML-r model,
there should be three OML layers in this model. 
Moreover, each OML layer is followed by a ReLU layer except the last OML layer (\emph{i.e.}, the third OML layer).
This principle is also satisfied for the MLOML-s and MLOML-t models.

A loss layer can also be added, which can give a global adjustment of the entire metric-algorithm-based model via backward propagation. 
To adequately use the effect of each local metric layer, the local loss is also utilized to update all the former layers (\emph{i.e.,} the loss of the $i$-th metric layer can be used to update the $1$-st to the $(i\!-\!1)$-th layers). 
In this way, vanishing gradient problem can also be alleviated. 
The novel loss function can be formulated as follows:
\begin{equation}\label{fun1} 
\Gamma=\frac{1}{2}\Gamma_{triplet}+\sum_{i=1}^{n}w_i\Gamma_{local}^i+\frac{\lambda}{2}\sum_{i=1}^n\lVert\bm{L}^i\rVert_F^2,
\end{equation}
where
$\Gamma_{triplet}=[\lVert\bm{x}_t^{(n)}-\bm{x}_p^{(n)}\rVert_2^2+1-\lVert\bm{x}_t^{(n)}-\bm{x}_q^{(n)}\rVert_2^2]_+$ indicates the triplet loss of the final output of the model (where $[z]_+=\max(0,z)$), $\Gamma_{local}^i$ denotes the local loss of the $i$-th OML layer (\emph{i.e.,} Eq.(\ref{fun4})), and $\lVert\bm{L}^i\rVert_F^2$ represents the Frobenius norm of parameter matrix $\bm{L}^i$, \emph{i.e.,} the transformation matrix learnt in the $i$-th OML layer. 
Moreover, $\lambda$ is the predefined hyper-parameter. 
While $w_i$, the weight of the $i$-th metric layer can be learnt by SGD during training phase, indicating the importance of each metric layer.

\paraheading{A New Mahalanobis-based OML (MOML):}
\label{MOML}
To build a metric layer of the proposed MLOML, a new OML algorithm named MOML is presented, which can act as a representative of Mahalanobis-based algorithms. 
Note that, in essence, MLOML can be constructed by other Mahalanobis-based algorithms. 
However, with MOML as a building component, MLOML enjoys better theoretical properties. 
The goal of MOML, learnt from triplet constraints, is to learn a Mahalanobis distance function $D$ that satisfies the following large margin constraint:
\begin{equation}\label{fun2} 
  D_{\bm{M}}(\bm{x},\bm{x}_q)>D_{\bm{M}}(\bm{x},\bm{x}_p)+r, \forall \bm{x},\bm{x}_p,\bm{x}_q\in\mathbb{R}^d,
\end{equation}
where $\bm{x}$ and $\bm{x}_p$ belong to the same class, while $\bm{x}$ and $\bm{x}_q$ come from different classes. $D_{\bm{M}}(\bm{x}_1,\bm{x}_2)=(\bm{x}_1-\bm{x}_2)^\top\bm{M}(\bm{x}_1-\bm{x}_2)$, where $\bm{M}\!\in\!\mathbb{R}^{d\times d}$ is a positive semi-definite parameter matrix. Also, $r$ is the margin. Naturally, the hinge loss (\emph{i.e.,} $r=1$) can be employed as below,
\begin{equation}\label{fun3} 
 \ell(\bm{M},\langle\bm{x},\bm{x}_p,\bm{x}_q\rangle)\!=\!\max(0,1\!+\!D_{\bm{M}}(\bm{x},\bm{x}_p)\!-\!D_{\bm{M}}(\bm{x},\bm{x}_q))\,.
\end{equation}

In a sequential manner, given a triplet $\langle\bm{x}_t,\bm{x}_p,\bm{x}_q\rangle$ at the $t$-th time step. Inspired by the Passive-Aggressive (PA) algorithms (\emph{i.e.,} a family of margin based online learning algorithms)~\cite{crammer2006online}, we design a convex objective function at each time step as follows,
\begin{equation}\label{fun4}\small
\begin{split}
      \Gamma&=\underset{\bm{M}\succcurlyeq0}{\arg\min}\frac{1}{2}\lVert\bm{M}-\bm{M}_{t-1}\rVert_F^2+\gamma\Big[1+D_{\bm{M}}(\bm{x},\bm{x}_p)\!-\!D_{\bm{M}}(\bm{x},\bm{x}_q)\Big]_+ \\
      &\scriptsize\text{\big($D_{\bm{M}}(\bm{x},\bm{x}_p)\!=\!(\bm{x}\!-\!\bm{x}_p)^\top\bm{M}(\bm{x}\!-\!\bm{x}_p)$, $D_{\bm{M}}(\bm{x},\bm{x}_q)\!=\!(\bm{x}\!-\!\bm{x}_q)^\top\bm{M}(\bm{x}\!-\!\bm{x}_q)$\big)}\\
      &=\underset{\bm{M}\succcurlyeq0}{\arg\min}\frac{1}{2}\lVert\bm{M}-\bm{M}_{t-1}\rVert_F^2+\gamma\Big[1+\Tr(\bm{M}\bm{A}_t)\Big]_+\,,
\end{split}
\end{equation}
where $\|\cdot\|_F$ is Frobenius norm, $[z]_+=\max(0,z)$ is the hinge loss, $\Tr(\cdot)$ denotes the trace operation, $\gamma$ is the regularization parameter and $\bm{A}_t\!=\!(\bm{x}_t\!-\!\bm{x}_p)(\bm{x}_t\!-\!\bm{x}_p)^\top\!-\!(\bm{x}_t\!-\!\bm{x}_q)(\bm{x}_t\!-\!\bm{x}_q)^\top$. We can easily get that $\Gamma$ is a convex function for $\bm{M}$, because $\Tr(\bm{M}\bm{A}_t)$ is a linear function of $\bm{M}$ which is convex, the hinge loss function $[1\!+\!z]_+$ is convex (not continuous at $z\!=\!-1$), and $\|\cdot\|_F$ and the domain $\bm{M}\!\succcurlyeq\!0$ are convex too. It can be shown that an optimal solution can be found within the domain $\bm{M}\!\succcurlyeq\!0$ by properly setting the value of $\gamma$. Thus, we can get the optimal solution of Eq.~(\ref{fun4}) by calculating the gradient $\frac{\partial\Gamma(\bm{M})}{\partial\bm{M}}\!=\!0$:
\begin{equation}\label{fun5} 
 \begin{split}
   &\frac{\partial\Gamma(\bm{M})}{\partial\bm{M}}=\left\{
   \begin{array}{ll}
   \bm{M}-\bm{M}_{t-1}+\gamma\bm{A}_t=0 & [z]_+>0\\
   \bm{M}-\bm{M}_{t-1}=0 & [z]_+=0\,.
   \end{array}
   \right.\\
   &s.t.\qquad \bm{M}\succcurlyeq0
 \end{split}
\end{equation}
According to Theorem~\ref{theorem1} (presented below), with a proper $\gamma$, the semi-positive definitiveness of $\bm{M}$ can be guaranteed. Thus, at the $t$-th time step, the parameter matrix $\bm{M}_t$ can be updated as below,
\begin{equation}\label{fun6} 
   \bm{M}_t=\left\{
   \begin{array}{ll}
   \bm{M}_{t-1}-\gamma\bm{A}_t & [z]_+>0\\
   \bm{M}_{t-1} & [z]_+=0.
   \end{array}
   \right.
\end{equation}
From Eq.~(\ref{fun6}), we can see that the time complexity of MOML is $O(d^2)$ at each time step. Using MOML as the base metric layer of MLOML has the following advantages: (1) the objective function of MOML is convex and enjoys a closed-form solution, which is beneficial to theoretical analysis; (2) without loss of generality, MOML can act as a representative of Mahalanobis-based OML algorithms.

\paraheading{Theoretical Guarantee:}
\label{theory}
Several theoretical guarantees are given for the proposed algorithms. Theorem~\ref{theorem1} is a positive-definite guarantee of the parameter matrix $\bm{M}$ in MOML. Moreover, Theorem~\ref{theorem2} presents a regret bound of MOML. Proposition~\ref{proposition1} tries to analyze and explain the effectiveness of the proposed framework \emph{i.e.,} MLOML. All the details of the proofs can be found in the appendix.

\begin{theorem}\label{theorem1}
Suppose $\bm{M}_t$ is positive-definite, then $\bm{M}_{t+1}$ given by the MOML update, i.e., $\bm{M}_{t+1}\!=\!\bm{M}_t\!-\!\gamma\bm{A}_{t+1}$ is positive definite by properly setting $\gamma$.
\end{theorem}

\begin{theorem}
Let $\langle\bm{x}_1, \bm{x}_p, \bm{x}_q \rangle,\cdots,\langle\bm{x}_T, \bm{x}_p, \bm{x}_q \rangle$ be a sequence of triplet constraints where each sample $\bm{x}_t|_{t=1}^T\in\mathbb{R}^d$ has $\|\bm{x}_t\|_2\!=\!1$ for all $t$. Let $\bm{M}_t\in\mathbb{R}^{d\times d}$ be the solution of MOML at the $t$-th time step, and $\bm{U}\in\mathbb{R}^{d\times d}$ denotes an arbitrary parameter matrix. By setting $\gamma\!=\!\frac{1}{\Phi\sqrt{T}}$ (where $\Phi\!\in\!\mathbb{R}^+$), the regret bound is
\begin{equation}\label{fun7}
  R(\bm{U},T)\!=\!\sum_{t=1}^T\ell(\bm{M}_t)\!-\!\sum_{t=1}^T\ell(\bm{U})\leq\frac{1}{2}\lVert\bm{I}\!-\!\bm{U}\rVert_F^2\!+\!\frac{32}{\Phi^2}\,.
\end{equation}
\label{theorem2}
\end{theorem}

\begin{proposition}
Let $\bm{M}_1,\cdots,\bm{M}_n$ be the parameter matrixes learnt by each metric layer of MLOML. The subsequent metric layer can learn a feature space that is at least as good as the one learnt by the former metric layer. That is, the composite feature space learnt by both $\bm{M}_1$ and $\bm{M}_2$ is better than the feature space learnt only by $\bm{M}_1$ in most cases (i.e., the feature space is more discriminating for classification).
\label{proposition1}
\end{proposition}

\paraheading{Other OML Algorithms}
In addition to MOML, other OML algorithms such as LEGO~\cite{jain2009online}, RDML~\cite{jin2009regularized} and OPML~\cite{li2017opml} \emph{etc.,} can also be adapted into the proposed multi-layer framework (namely LEGO-Multi, RDML-Multi and OPML-Multi). 
It is worth mentioning that both LEGO and RDML learn a Mahalanobis parameter matrix $\bm{M}$, while OPML just learns a transformation matrix $\bm{L}$. 
Hence, OPML doesn't need an additional matrix decomposition operation $(\emph{i.e.,}\;\bm{M}=\bm{L}^\top\bm{L})$. 
The experimental results of LEGO-Multi, RDML-Multi and OPML-Multi will be discussed in Section~\ref{other-OML}.

\subsection{Forward and Backward Propagation}
The proposed MLOML (\emph{i.e.,} MLOML-r, MLOML-s, MLOML-t) is made up of a series of OML algorithms (\emph{i.e.,} MOML metric layer) and nonlinear functions (\emph{i.e.,} ReLU, Sigmoid, tanh).
Then, MLOML attempts to explore a new way to train the metric layer by introducing forward propagation (FP) updating. 
In fact, MLOML can not only be learnt by forward propagation, but also be learnt by backward propagation. 
Moreover, these two strategies can be adopted simultaneously too. 
During forward propagation, each metric layer can be learnt immediately, through this way, new feature space can be explored sequentially. 
When backward propagation, the return gradients can be used to fine-tune all the metric layers, amending the feature spaces learnt by the forward propagation.

\begin{figure}[!tbp]
  \centering
  \includegraphics[width=0.48\textwidth]{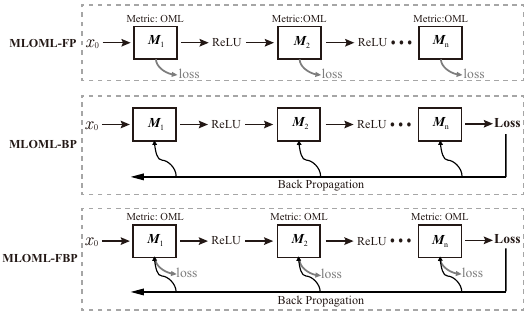}
  \caption{Flowcharts of MLOML-FP, MLOML-BP and MLOML-FBP, respectively, where MLOML is the MLOML-r model}.
  \label{fig-ODML-FBP}
\end{figure}

Therefore, MLOML can be trained with three different propagation strategies as follows: (1) \textbf{MLOML-FP}, which is only trained by employing forward propagation strategy. (2) \textbf{MLOML-FBP}, which utilizes forward and backward propagation strategies simultaneously. Specifically, a loss layer is added as the last layer to calculate the final loss, where the loss function (\emph{i.e.,} Eq.~(\ref{fun1})) is adopted. (3) \textbf{MLOML-BP} is similar to MLOML-FBP, while MLOML-BP only utilizes the final loss to train the entire model without the local losses and without forward updating. 
The flowcharts of these three variations can be seen in Fig.~\ref{fig-ODML-FBP}. 
The comparison between these variations will be shown in Section~\ref{FP-BP-FBP}.

\begin{table}[!tbp] \small
\centering
\extrarowheight=2.4pt
\tabcolsep=3.3pt
\caption{Twelve UCI datasets with different scales (\emph{i.e.,} \#inst) and feature dimensions (\emph{i.e.,} \#feat).}
\begin{tabular}{lrrr|lrrr}
  \hline
  Datasets     &\#inst  &\#feat &\#class  &Datasets    &\#inst   &\#feat &\#class\\
  \hline
  lsvt         &126     &310    &2        &balance     &625      &4      &3      \\
  iris         &150     &4      &3        &breast      &683      &9      &2      \\
  wine         &178     &13     &3        &pima        &768      &8      &2      \\
  spect        &267     &22     &2        &diabetic    &1151     &19     &2      \\
  ionophere    &351     &34     &2        &waveform    &5000     &21     &3      \\
  pems         &440     &137710 &7        &mlprove     &6118     &57     &6      \\
  \hline
\end{tabular}
\label{Uci-dataset}
\end{table}

\begin{table*}[!tbp] 
\centering
\caption{Error rates (mean $\pm$ std. deviation) on the UCI datasets. $\bullet/\circ$ indicates that MLOML-r, MLOML-s, MLOML-t are significantly better/worse than the respective algorithm according to the $t$-tests at $95\%$ significance level. The statistics of win/tie/loss between MLOML-r and other algorithms is also counted.}
\resizebox{\textwidth}{!}{
    \begin{tabular}{l|ccccccc}
    \toprule
    \multirow{2}[4]{*}{Datasets} & \multicolumn{1}{c|}{\multirow{2}[4]{*}{Euclidean}} & \multicolumn{4}{c|}{Batch} & \multicolumn{2}{c}{Online} \\
\cmidrule{3-8}    \multicolumn{1}{c|}{} & \multicolumn{1}{c|}{} & LMNN  & KISSME & LMDML & \multicolumn{1}{c|}{ML-CC} & RDML  & LEGO \\
    \midrule
    Isvt  & .369$\pm$.051$\bullet$ & .387$\pm$.057$\bullet$ & .403$\pm$.098$\bullet$ & .374$\pm$.064$\bullet$ & .384$\pm$.047$\bullet$ & .400$\pm$.055$\bullet$ & {.369$\pm$.051$\bullet$} \\
    iris  & .038$\pm$.016$\bullet$ & .040$\pm$.018$\bullet$ & .039$\pm$.020$\bullet$ & .028$\pm$.015 & .058$\pm$.034$\bullet$ & .028$\pm$.019 & {.037$\pm$.016$\bullet$} \\
    wine  & .218$\pm$.039 & .170$\pm$.044$\circ$ & \textbf{.069$\pm$.021$\circ$} & .229$\pm$.041 & .112$\pm$.031$\circ$ & .350$\pm$.028$\bullet$ & {.231$\pm$.041$\bullet$} \\
    spect & .354$\pm$.031$\bullet$ & .357$\pm$.032$\bullet$ & .365$\pm$.031$\bullet$ & .342$\pm$.036$\bullet$ & .409$\pm$.044$\bullet$ & .347$\pm$.035$\bullet$ & {.326$\pm$.035} \\
    ionophere & .180$\pm$.017$\bullet$ & .157$\pm$.016$\bullet$ & .156$\pm$.023$\bullet$ & .115$\pm$.012$\bullet$ & .104$\pm$.022$\bullet$ & .096$\pm$.015$\bullet$ & {.129$\pm$.019$\bullet$} \\
    pems  & .498$\pm$.033$\bullet$ & .402$\pm$.038 & \textbf{.188$\pm$.028$\circ$} & .352$\pm$.028$\circ$ & .227$\pm$.025$\circ$ & .421$\pm$.030$\bullet$ & {.461$\pm$.033$\bullet$} \\
    balance & .108$\pm$.013$\bullet$ & .088$\pm$.013$\bullet$ & .101$\pm$.011$\bullet$ & .075$\pm$.010$\bullet$ & .066$\pm$.011 & .070$\pm$.011 & {.091$\pm$.011$\bullet$} \\
    breast & .106$\pm$.012 & .107$\pm$.012 & .106$\pm$.014 & .107$\pm$.013 & .113$\pm$.012$\bullet$ & .115$\pm$.015$\bullet$ & {\textbf{.104$\pm$.016}} \\
    pima  & .324$\pm$.018 & .326$\pm$.020 & .333$\pm$.021$\bullet$ & .330$\pm$.019$\bullet$ & .322$\pm$.022 & .357$\pm$.022$\bullet$ & {.322$\pm$.020} \\
    diabetic & .343$\pm$.018 & .335$\pm$.017$\circ$ & \textbf{.288$\pm$.018$\circ$} & .316$\pm$.014$\circ$ & .338$\pm$.011 & .353$\pm$.015$\bullet$ & {.322$\pm$.006$\circ$} \\
    waveform & .195$\pm$.006$\bullet$ & .190$\pm$.005$\bullet$ & \textbf{.158$\pm$.006$\circ$} & .176$\pm$.007$\circ$ & .208$\pm$.006$\bullet$ & .166$\pm$.006$\circ$ & {.198$\pm$.006$\bullet$} \\
    mlprove & .084$\pm$.005$\bullet$ & .037$\pm$.004$\bullet$ & .234$\pm$.273$\bullet$ & .007$\pm$.002$\bullet$ & .032$\pm$.025$\bullet$ & .027$\pm$.011$\bullet$ & {.024$\pm$.003$\bullet$} \\
    \midrule
    \textbf{win/tie/loss} & \textbf{8/4/0} & \textbf{7/3/2} & \textbf{7/1/4} & \textbf{6/3/3} & \textbf{7/3/2} & \textbf{9/2/1} & \textbf{8/3/1} \\
    \midrule
    \midrule
    \multirow{2}[4]{*}{Datasets} & \multicolumn{7}{c}{Online} \\
\cmidrule{2-8}    \multicolumn{1}{c|}{} & OASIS & OPML  & SLMOML & MOML  & MLOML-t & MLOML-s & MLOML-r \\
    \midrule
    Isvt  & .333$\pm$.000 & .370$\pm$.051$\bullet$ & .369$\pm$.051$\bullet$ & .369$\pm$.051$\bullet$ & .369$\pm$.054$\bullet$ & .369$\pm$.052$\bullet$ & {\textbf{.326$\pm$.053}} \\
    iris  & .333$\pm$.000$\bullet$ & .035$\pm$.016$\bullet$ & .038$\pm$.016$\bullet$ & .028$\pm$.018 & \textbf{.027$\pm$.015} & \textbf{.025$\pm$.017} & {\textbf{.026$\pm$.017}} \\
    wine  & .586$\pm$.061$\bullet$ & .220$\pm$.040 & .225$\pm$.040 & .226$\pm$.041 & .214$\pm$.038$\circ$ & .216$\pm$.039 & {.219$\pm$.039} \\
    spect & .385$\pm$.033$\bullet$ & .321$\pm$.029 & .355$\pm$.032$\bullet$ & .331$\pm$.032$\bullet$ & .323$\pm$.028 & \textbf{.317$\pm$.024} & {\textbf{.320$\pm$.025}} \\
    ionophere & .183$\pm$.017$\bullet$ & .107$\pm$.020$\bullet$ & .107$\pm$.020$\bullet$ & .108$\pm$.032$\bullet$ & .102$\pm$.021$\bullet$ & \textbf{.086$\pm$.013$\bullet$} & {\textbf{.081$\pm$.016}} \\
    pems  & .651$\pm$.036$\bullet$ & .331$\pm$.029$\circ$ & .495$\pm$.033$\bullet$ & .416$\pm$.057$\bullet$ & .319$\pm$.032$\circ$ & .407$\pm$.030 & {.397$\pm$.036} \\
    balance & .125$\pm$.010$\bullet$ & .073$\pm$.012$\bullet$ & .077$\pm$.012$\bullet$ & .070$\pm$.010$\bullet$ & \textbf{.064$\pm$.013} & .066$\pm$.011 & {.066$\pm$.012} \\
    breast & .175$\pm$.050$\bullet$ & .109$\pm$.014$\bullet$ & .106$\pm$.012 & .112$\pm$.014$\bullet$ & .106$\pm$.013 & \textbf{.104$\pm$.013} & {.105$\pm$.012} \\
    pima  & .349$\pm$.003$\bullet$ & .324$\pm$.022 & .334$\pm$.022$\bullet$ & .323$\pm$.020 & .322$\pm$.019 & \textbf{.321$\pm$.017} & {.323$\pm$.018} \\
    diabetic & .451$\pm$.021$\bullet$ & .322$\pm$.019$\circ$ & .348$\pm$.016$\bullet$ & .342$\pm$.016 & .341$\pm$.015 & .341$\pm$.014 & {.342$\pm$.017} \\
    waveform & .298$\pm$.049$\bullet$ & .175$\pm$.006$\circ$ & .161$\pm$.005$\circ$ & .173$\pm$.006$\circ$ & .168$\pm$.006$\circ$ & .162$\pm$.005$\circ$ & {.187$\pm$.006} \\
    mlprove & .002$\pm$.001$\circ$ & .006$\pm$.002$\bullet$ & .011$\pm$.004$\bullet$ & \textbf{.001$\pm$.001$\circ$} & .002$\pm$.001$\circ$ & .002$\pm$.001$\circ$ & .004$\pm$.001 \\
    \midrule
    \textbf{win/tie/loss} & \textbf{10/1/1} & \textbf{6/3/3} & \textbf{9/2/1} & \textbf{6/4/2} & \textbf{2/6/4} & \textbf{2/8/2} &  \\
    \bottomrule
    \end{tabular}%
    }
\label{Uci-Results}
\end{table*}

\section{Experiments}
To verify the effectiveness and applicability of the proposed MLOML, we conduct various experiments on the UCI datasets, which include multiple real-world machine learning tasks for which only vectorized features can be accessed, to analyze and interpret the properties of MLOML.
First, we introduce the training process of MLOML.

\subsection{Training}
We will describe how to train MLOML in detail in this section. 
Note that, MLOML is trained from scratch in an end-to-end manner, which is totally different from the traditional layer-by-layer training.

\textbf{Initialization:} Parameter matrix $\bm{M}_i~(i\!=\!1,2,\ldots,n)$ is initialized as an identity matrix. The hyper-parameter $\gamma$ in MOML and the $\lambda$ in loss layer need to be chosen by cross-validation according to the specific task. All $w_i~(i=1, 2, \ldots, n)$ is initialized as $1$. The number of layers in MLOML is also a hyper-parameter, which can be chosen according to a specific task ($3$ or $5$ layers are usually enough).

\textbf{Forward Propagation:} At the $t$-th time step, one triplet $\langle\bm{x}_t^{(0)},\bm{x}_p^{(0)},\bm{x}_q^{(0)}\rangle$ is constructed. Then the triplet is fed into the first OML layer, and the current local triplet loss (\emph{i.e.,} Eq.~(\ref{fun4})) is calculated by using the current metric matrix $\bm{M}_1$. 
According to the updating strategy of MOML (\emph{i.e.,} Eq.~(\ref{fun6})), the metric matrix $\bm{M}_1$ is updated for the first time. 
Then, $\bm{M}_1$ is mathematically decomposed as $\bm{L}_1^\top\bm{L}_1$. 
After transformation by using $\bm{L}_1$, the new triplet $\langle\bm{x}_t^{(1)}\!=\!\bm{L}_1\bm{x}_t^{(0)}, \bm{x}_p^{(1)}\!=\!\bm{L}_1\bm{x}_p^{(0)}, \bm{x}_q^{(1)}\!=\!\bm{L}_1\bm{x}_q^{(0)}\rangle$ is fed into the next ReLU, Sigmoid or tanh layer. 
In a serial manner, the final output of the last layer is $\langle\bm{x}_t^{(n)}, \bm{x}_p^{(n)}, \bm{x}_q^{(n)}\rangle$. 
Through the linear (\emph{i.e.}, OML layer) and nonlinear transformation (\emph{i.e.}, ReLU, Sigmoid, or tanh layer), new feature spaces are sequentially explored. At the same time, the metric matrix of each OML layer is also learnt.

\textbf{Backward Propagation (optional):} The final loss is calculated according to Eq.~(\ref{fun1}) by using the output of the last OML layer. 
By using chain rule, SGD is adopted to update all the decomposed transformation matrix $\bm{L}_i~(i=1, 2, \ldots, n)$. 
Then each metric matrix $\bm{M}_i~(i=1, 2, \ldots, n)$ can be obtained naturally by $\bm{M}_i=\bm{L}_i^\top\bm{L}_i$. 
Note that all these three samples in a triplet are used to calculate the gradients. 
Ideally, forward updating can explore new feature spaces, while backward updating can amend the exploration. 
In this way, that is, exploration with amendment, a much better feature space can be found. 
In practice, the backward propagation indeed can further slightly improve the feature space learnt by the forward propagation in some cases, but this could also bring additional computation load. 
As a trade-off between time and performance, if not specified, we will train the proposed MLOML only by forward propagation, similar to the Deep forest~\cite{ZhouF17}. More details can be seen in Section~\ref{FP-BP-FBP}.

\subsection{Datasets}
We pick twelve commonly used datasets from UCI Machine Learning Repository~\cite{Lichman:2013}, which vary in the dimensionality and size. The details of these twelve datasets can be seen in Table~\ref{Uci-dataset}. The reason of choosing these datasets is that they are all vectorized data and can be representative data for the real-world applications. For example, \textit{lsvt} is a real voice rehabilitation treatment dataset. \textit{pems} contains $15$ months worth of daily data that describes the occupancy rate of different car lanes of the San Francisco bay area freeways. Also, \textit{ionophere} is real radar data, which is collected by a system in Goose Bay, Labrador.

Classification task will be conducted on these datasets. For each dataset, $50\%$ samples are randomly sampled as training set, and the rest is taken as test set. Each dataset will be resampled $30$ times, and each algorithm will be tested on all these sampled datasets. When the feature dimensionality $d\!\ge\!200$, the $d$-dimensional feature will be reduced to a $100$-dimensional feature by principal component analysis (PCA) for easier handling. All datasets are normalized by $\ell_2$ normalization. Error rate is adopted as the evaluation criterion.

\subsection{Comparison with the State of the Art}
To evaluate the effectiveness of the family of MLOML (MLOML-r, MLOML-s and MLOML-t), six state-of-the-art online metric learning (OML) algorithms, \emph{i.e.,} RDML~\cite{jin2009regularized}, LEGO~\cite{jain2009online}, OASIS~\cite{chechik2010large}, OPML~\cite{li2017opml}, SLMOML~\cite{zhong2018slmoml} and the new designed MOML are employed as comparisons. Note that all of these compared OML algorithms are single layer algorithms, while the proposed famliy of MLOML (MLOML-r, MLOML-s and MLOML-t) are built on MOML is multi-layer algorithm. Euclidean distance is adopted as the baseline algorithm. 
Besides, four batch metric learning algorithms \emph{i.e.,} LMNN~\cite{weinberger2005distance}, KISSME~\cite{koestinger2012large}, LMDML~\cite{nguyen2020scalable} and ML-CC~\cite{guo2021metric} are also employed for reference. 
Note that these three algorithms (\emph{i.e.,} Euclidean, LMNN and KISSME) are offline. Cross-validation is used for hyper-parameter selection for all algorithms. 
Specifically, the regularization parameter $\gamma$ for the family of MLOML (\emph{i.e.,} the $\gamma$ in MOML metric layer, $\gamma\in\{10^{-4}, 10^{-3}, 10^{-2}, 10^{-1}\}$), the learning rate $\lambda$ for RDML ($\lambda\in\{10^{-4}, 10^{-3}, 10^{-2}, 10^{-1}\}$), the regularization parameter $\eta$ for LEGO ($\eta\in\{10^{-4}, 10^{-3}, 10^{-2}, 10^{-1}\}$), the regularization parameter $\gamma$ for OPML ($\gamma\in\{10^{-4}, 10^{-3}, 10^{-2}, 10^{-1}\}$), the weighting parameter $\mu$ for LMNN ($\mu\in\{0.125, 0.25, 0.5\}$), the parameters $K$ and $\mu$ for ML-CC ($K\in\{2,4,8\}$ and $\mu\in\{0.1,0.3,0.5,0.7,0.9\}$, and the aggressiveness parameter $C$ for SLMOML ($C\in\{10^{-4}, 10^{-3}, 10^{-2}, 10^{-1}\}$) are all set up in this way.

\begin{table*}[!htp]
  \centering
  \caption{Error rates on twelve UCI datasets by employing different propagation strategies for MLOML.}
  \resizebox{\textwidth}{!}{
    \begin{tabular}{l|ccc|ccc|ccc}
    \toprule
    \multirow{2}[4]{*}{Datasets} & \multicolumn{3}{c|}{MLOML-r} & \multicolumn{3}{c|}{MLOML-s} & \multicolumn{3}{c}{MLOML-t} \\
\cmidrule{2-10}    \multicolumn{1}{c|}{} & BP    & FBP   & \multicolumn{1}{c|}{FP} & BP    & FBP   & \multicolumn{1}{c|}{FP} & BP    & FBP   & \multicolumn{1}{c}{FP} \\
    \midrule
    Isvt  & .354$\pm$.053$\bullet$ & .325$\pm$.055 & \multicolumn{1}{p{4.04em}|}{.326$\pm$.053} & .369$\pm$.051 & .369$\pm$.052 & \multicolumn{1}{p{4.04em}|}{.369$\pm$.052} & .369$\pm$.051 & .369$\pm$.051 & {.369$\pm$.054} \\
    iris  & .032$\pm$.015$\bullet$ & .026$\pm$.017 & \multicolumn{1}{p{4.04em}|}{.026$\pm$.017} & .040$\pm$.017$\bullet$ & .025$\pm$.017 & \multicolumn{1}{p{4.04em}|}{.025$\pm$.017} & .039$\pm$.019$\bullet$ & .027$\pm$.015 & {.027$\pm$.015} \\
    wine  & .220$\pm$.040 & .220$\pm$.040 & \multicolumn{1}{p{4.04em}|}{.219$\pm$.039} & .216$\pm$.039 & .216$\pm$.039 & \multicolumn{1}{p{4.04em}|}{.216$\pm$.039} & .216$\pm$.041 & .214$\pm$.041 & {.214$\pm$.038} \\
    spect & .358$\pm$.029$\bullet$ & .319$\pm$.026 & \multicolumn{1}{p{4.04em}|}{.320$\pm$.025} & .352$\pm$.030$\bullet$ & .314$\pm$.025$\bullet$ & \multicolumn{1}{p{4.04em}|}{.317$\pm$.024} & .346$\pm$.030$\bullet$ & .321$\pm$.025 & {.323$\pm$.028} \\
    ionophere & .128$\pm$017$\bullet$ & .081$\pm$.017 & \multicolumn{1}{p{4.04em}|}{.081$\pm$.016} & .175$\pm$.016$\bullet$ & .088$\pm$.013 & \multicolumn{1}{p{4.04em}|}{.086$\pm$.013} & .150$\pm$.016$\bullet$ & .102$\pm$.021 & {.102$\pm$.021} \\
    pems  & .466$\pm$.036$\bullet$ & .396$\pm$.033 & \multicolumn{1}{p{4.04em}|}{.397$\pm$.036} & .500$\pm$.032$\bullet$ & .408$\pm$.030 & \multicolumn{1}{p{4.04em}|}{.407$\pm$.030} & .496$\pm$.032$\bullet$ & .319$\pm$.031 & {.319$\pm$.032} \\
    balance & .070$\pm$.013$\bullet$ & .066$\pm$.011 & \multicolumn{1}{p{4.04em}|}{.066$\pm$.012} & .109$\pm$.012$\bullet$ & .066$\pm$.012 & \multicolumn{1}{p{4.04em}|}{.066$\pm$.011} & .066$\pm$.011$\bullet$ & .064$\pm$.013 & {.064$\pm$.013} \\
    breast & .109$\pm$.015$\bullet$ & .107$\pm$.014 & \multicolumn{1}{p{4.04em}|}{.105$\pm$.012} & .106$\pm$.012 & .104$\pm$.013 & \multicolumn{1}{p{4.04em}|}{.104$\pm$.013} & .106$\pm$.016 & .104$\pm$.014 & {.106$\pm$.013} \\
    pima  & .323$\pm$.017 & .324$\pm$.017 & \multicolumn{1}{p{4.04em}|}{.323$\pm$.018} & .322$\pm$.017 & .321$\pm$.017 & \multicolumn{1}{p{4.04em}|}{.321$\pm$.017} & .323$\pm$.018 & .323$\pm$.020 & \multicolumn{1}{p{4.04em}}{.322$\pm$.019} \\
    diabetic & .340$\pm$.017 & .340$\pm$.014 & \multicolumn{1}{p{4.04em}|}{.342$\pm$.017} & .342$\pm$.016 & .340$\pm$.014 & \multicolumn{1}{p{4.04em}|}{.341$\pm$.014} & .341$\pm$.015 & .342$\pm$.015 & {.341$\pm$.015} \\
    waveform & .180$\pm$.006$\circ$ & .175$\pm$.006$\circ$ & \multicolumn{1}{p{4.04em}|}{.187$\pm$.006} & .194$\pm$.006$\bullet$ & .163$\pm$.004 & \multicolumn{1}{p{4.04em}|}{.162$\pm$.005} & .169$\pm$.005 & .166$\pm$.005$\bullet$ & {.168$\pm$.006} \\
    mlprove & .006$\pm$.002$\bullet$ & .003$\pm$.001$\circ$ & \multicolumn{1}{p{4.04em}|}{.004$\pm$.001} & .084$\pm$.005$\bullet$ & .002$\pm$.001 & \multicolumn{1}{p{4.04em}|}{.002$\pm$.001} & .007$\pm$.002$\bullet$ & .001$\pm$.001$\bullet$ & {.002$\pm$.001} \\
    \midrule
    \textbf{win/tie/loss} & \textbf{8/3/1} & \textbf{0/10/2} &       & \textbf{7/5/0} & \textbf{0/11/1} &       & \textbf{6/6/0} & \textbf{0/10/2} &  \\
    \bottomrule
    \end{tabular}%
    }
\label{Uci-ODML-FBP}
\end{table*}

For fair comparison, all OML algorithms adopt the same triplet construction strategy introduced by OPML to construct the pairwise or triplet constraints. The difference is that, in OPML the triplet construction strategy is one-pass, while here multiple-scan strategy is employed to construct more constraints for adequately training (the scanning number is set to $20$). Note that, all OML algorithms are still trained in an online manner. Moreover, three metric layers MLOML is adopted in this experiment. A $k$-NN classifier (\emph{i.e.,} $k = 5$) is used to get the final classification results. The results are summarized in Table~\ref{Uci-Results}. For each dataset, the mean and standard deviation of error rate are calculated, and pairwise $t$-tests between MLOML and other algorithms at $95\%$ significance level are also performed. Then the win/tie/loss is counted according to the $t$-test. From this table, we can see that the family of MLOML can not only achieve superior performance compared with other state-of-the-art OML algorithms, but also better than batch metric learning algorithms except KISSME algorithm.
A possible reason is that KISSME learns a distance metric from equivalence constraints which is easier to specify labels.
We can also see that MLOML is robust on small datasets, \emph{e.g.,} lsvt, iris, spect and ionophere, which means that MLOML can handle small-scale data very well.

\begin{figure}[!tbp]
  \centering
  \subfigure{\includegraphics[width=0.24\textwidth]{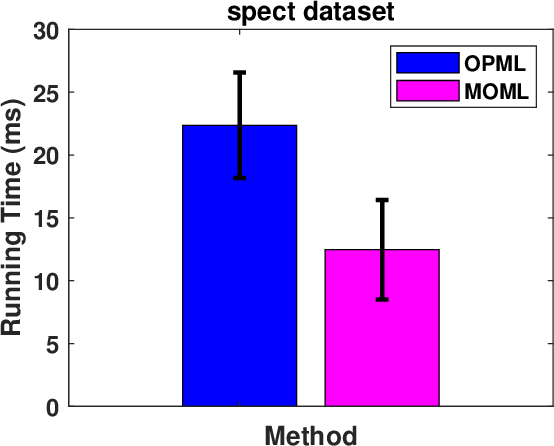}}
  \subfigure{\includegraphics[width=0.24\textwidth]{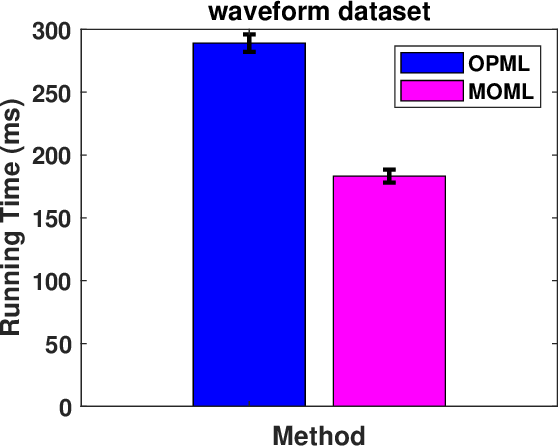}}
  \caption{\textcolor[rgb]{0.00,0.00,1.00}{Running time of OPML and MOML on the spect and waveform datasets.}} 
  \label{fig-MOML-OPML}
\end{figure}

\textcolor[rgb]{0.00,0.00,1.00}{The traditional online metric learning algorithms, RDML, LEGO, OASIS, OPML and SLMOML algorithms, like our proposed MOML algorithm, can be used as metric layers for multi-layer online metric learning algorithms. However, in the literature of the RDML, LEGO, OASIS and
SLMOML algorithms, they need to first sample triplets, and then apply these triplets to the training process. Because of this, their running time are relatively long. In order to guarantee the fairness of the comparison, Fig. \ref{fig-MOML-OPML} shows the running time of OPML and MOML on the spect and waveform datasets, where OPML and MOML use the one-pass triplet construction strategy. We can see that the proposed MOML has a shorter running time than OPML.}

\begin{figure*}[!tbp]
\centering
\subfigure{
           \includegraphics[width=0.24\textwidth]{Metric_layer_MOML/lsvt.eps}}
\subfigure{
           \includegraphics[width=0.24\textwidth]{Metric_layer_MOML/iris.eps}}
\subfigure{
           \includegraphics[width=0.24\textwidth]{Metric_layer_MOML/spect.eps}}
\subfigure{
           \includegraphics[width=0.24\textwidth]{Metric_layer_MOML/ionosphere.eps}}
\subfigure{
           \includegraphics[width=0.24\textwidth]{Metric_layer_MOML/pems.eps}}
\subfigure{
           \includegraphics[width=0.24\textwidth]{Metric_layer_MOML/balance.eps}}
\subfigure{
           \includegraphics[width=0.24\textwidth]{Metric_layer_MOML/wine.eps}}
\subfigure{
           \includegraphics[width=0.24\textwidth]{Metric_layer_MOML/breast.eps}}
\subfigure{
           \includegraphics[width=0.24\textwidth]{Metric_layer_MOML/mlprove.eps}}
\caption{\textcolor[rgb]{0.00,0.00,1.00}{Results of different metric layers of MLOML. Moreover, Euclidean, MOML and LMNN are taken as the baseline algorithms.}}
\label{fig-diff-layer}
\end{figure*}
\begin{figure*}[!tbp]
\centering
\subfigure{
           \includegraphics[width=0.62\textwidth]{feature_space2/balance.eps}}
\subfigure{
           \includegraphics[width=0.62\textwidth]{feature_space2/ionosphere.eps}}
\subfigure{
           \includegraphics[width=0.62\textwidth]{feature_space2/iris.eps}}
\subfigure{
           \includegraphics[width=0.62\textwidth]{feature_space2/mlprove.eps}}
\caption{ Feature visualization on four UCI datasets, demonstrating the feature representation learnt by each metric layer in MLOML-3L.}
\label{fig-FeatureSpace}
\end{figure*}

\begin{figure*}[!tp]
\centering
\subfigure{
           \includegraphics[width=0.24\textwidth]{Epoch_eps/pima.eps}}
\subfigure{
           \includegraphics[width=0.24\textwidth]{Epoch_eps/iris.eps}}
\subfigure{
           \includegraphics[width=0.24\textwidth]{Epoch_eps/diabetic.eps}}
\subfigure{
           \includegraphics[width=0.24\textwidth]{Epoch_eps/spect.eps}}
\subfigure{
           \includegraphics[width=0.24\textwidth]{Epoch_eps/pems.eps}}
\subfigure{
           \includegraphics[width=0.24\textwidth]{Epoch_eps/lsvt.eps}}
\subfigure{
           \includegraphics[width=0.24\textwidth]{Epoch_eps/ionosphere.eps}}
\subfigure{
           \includegraphics[width=0.24\textwidth]{Epoch_eps/balance.eps}}
\subfigure{
           \includegraphics[width=0.24\textwidth]{Epoch_eps/mlprove.eps}}
\caption{Error rates on nine UCI datasets by changing the number of scans for MOML and MLOML.}
\label{fig-change-epoch}
\end{figure*}

\begin{figure*}[!tbp]
\centering
\subfigure{
           \includegraphics[width=0.24\textwidth]{Metric_layer_OtherOML/lsvt.eps}}
\subfigure{
           \includegraphics[width=0.24\textwidth]{Metric_layer_OtherOML/iris.eps}}
\subfigure{
           \includegraphics[width=0.24\textwidth]{Metric_layer_OtherOML/spect.eps}}
\subfigure{
           \includegraphics[width=0.24\textwidth]{Metric_layer_OtherOML/ionosphere.eps}}
\subfigure{
           \includegraphics[width=0.24\textwidth]{Metric_layer_OtherOML/pems.eps}}
\subfigure{
           \includegraphics[width=0.24\textwidth]{Metric_layer_OtherOML/balance.eps}}
\subfigure{
           \includegraphics[width=0.24\textwidth]{Metric_layer_OtherOML/wine.eps}}
\subfigure{
           \includegraphics[width=0.24\textwidth]{Metric_layer_OtherOML/diabetic.eps}}
\subfigure{
           \includegraphics[width=0.24\textwidth]{Metric_layer_OtherOML/mlprove.eps}}
\caption{\textcolor[rgb]{0.00,0.00,1.00}{Results of different metric layers of RDML-multi, LEGO-multi and OPML-multi (marked by hollow shapes), which are stacked by online metric algorithms RDML, LEGO and OPML (marked by corresponding solid shapes) along with the ReLU layers, respectively.}}
\label{fig-diff-layer2}
\end{figure*}

\subsection{Forward and Backward Propagation}
\label{FP-BP-FBP}
In this section, we analyze the learning ability of MLOML by adopting different propagation strategies, \emph{i.e.,} MLOML-FP, MLOML-BP and MLOML-FBP. Specifically, we conduct classification task on the twelve UCI datasets to compare these three variations of MLOML, each of which contains three metric layers. The results are exhibited in Table~\ref{Uci-ODML-FBP}. From the results, we can see that MLOML-FP performs better than MLOML-BP. The reason is not difficult to perceive, because BP may suffer from the vanishing gradient problem. Taking advantage of the fact that each metric layer of MLOML is a MOML algorithm, it can learn a good metric in each layer during FP. We can also observe that MLOML-FP performs similarly to MLOML-FBP. The reason may be that MLOML-FP has achieved quite good classification performance on some datasets, so additional BP updating cannot further improve the performance. However, on other datasets, MLOML-FBP indeed achieves the best classification performance as expected, such as iris, spect and mlprove \emph{etc.} It is worth mentioning that MLOML-FP is the fastest one among these three variations with a time complexity of $O(nd^2)$, where $n$ is the number of metric layers. Overall, for the proposed MLOML, the FP training strategy is the best one when considering both training performance and training efficiency. 
It should be noted that in the following chapters, MLOML refers specifically to the MLOML-r model, and the three-layer and five-layer MLOML networks are denoted by MLOML-3L and MLOML-5L.

\subsection{Progressive Feature Representation}
\label{hier-fea}
In this section, we will analyze the progressive feature representation ability of each metric layer in MLOML and verify that the metric space can become better and better by adding metric layer gradually. Particularly, an MLOML-5L model is employed. To test the feature representation ability of each metric layer, we perform classification task on the output features of each metric layer respectively. We choose nine UCI datasets and take Euclidean distance, MOML and LMNN as the baseline algorithms. Note that only the test sets of these datasets are used to perform this experiment. \textcolor[rgb]{0.00,0.00,1.00}{From Fig.~\ref{fig-diff-layer}, we can see that the classification performance of MLOML-5L becomes better with the increase of the number of metric layers.} Besides, in some datasets, the curve of error rate can converge smoothly. 
\textcolor[rgb]{0.00,0.00,1.00}{Moreover, we apply PCA to four UCI datasets to obtain the new space with 2-dimensional features, and then visualize the feature space learnt by each metric layer for more intuition (shown in Fig.~\ref{fig-FeatureSpace}).}
The four UCI datasets are picked and entered into one learnt MLOML-3L model. 
Next, all output samples of each metric layer are $\ell_2$ normalized and reduced to a two-dimensional space by PCA. As seen, in original feature space, the distribution of samples is disordered. As the number of metric layers increases, the intra-class distance becomes smaller, the inter-class distance becomes larger, and the distribution of samples becomes more separable.

\subsection{Learning Ability of MLOML}
Since the multiple-scan strategy is performed in the training phase, it is necessary to test the learning ability of MLOML by setting different numbers of scans. Note that $m$ times scanning will scan the training data $m$ times. Therefore, we set the number of scans from $1$ to $20$, and compare the classification performance between MLOML and MOML under different scans. Specifically, nine datasets are picked, and Euclidean distance is taken as the baseline. The results are presented in Fig.~\ref{fig-change-epoch}. From the figure, we can see that as the number of scans increases, the classification performance of MLOML is significantly improved and then converge, which can reflect the ability of MLOML for reusing data. Compared with MOML, with the same amount of data (\emph{i.e.,} the same scan), MLOML can learn better feature representation (\emph{i.e.,} lower error rate). In other words, the learning ability of MLOML is stronger than MOML, which means that MLOML can gain more learning ability from the multi-layer architecture.

\subsection{Extendability of MLOML}
\label{other-OML}
In order to verify the extendability of the proposed framework, we take the other three OML algorithms (\emph{e.g.,} LEGO, RDML and OPML) as the base OML layer followed by the ReLU layer and construct their corresponding multi-layer versions, respectively (\emph{i.e.,} LEGO-multi, RDML-multi and OPML-multi). Note that these three algorithms are all Mahalanobis-based OML algorithms. For simplicity, FP strategy is employed for these three algorithms. Other settings are similar to the ones in Section~\ref{hier-fea}. 
\textcolor[rgb]{0.00,0.00,1.00}{From Fig.~\ref{fig-diff-layer2}, we can see that LEGO-multi, RDML-multi and OPML-multi have similar characteristic to MLOML}. In most cases, multi-layer versions of these algorithms perform better than their corresponding shallow versions. Moreover, the progressive learning ability of feature representation is demonstrated. Therefore, the effectiveness and extendability of the proposed framework are verified.

\section{Discussions and Conclusions}
In this study, we propose a multi-layer framework for online metric learning. 
Specifically, we implement \textit{multi-layer online metric learning (MLOML)} by stacking a set of OML algorithms. Extensive experiments have been conducted to analyze and verify the properties of MLOML. For future work, we will analyze and discuss this framework from three aspects as follows.
\begin{itemize}
    \item \textbf{Extendability:} Although only OML-based algorithms are implemented (\emph{e.g.,} MLOML), the proposed framework is extensible, such as: a) mini-batch or batch metric learning based metric layer can be constructed; b) different metric learning algorithms can be combined as different metric layers.
    \item \textbf{Advantages:} The proposed MLOML has many nice properties: a) it is online; b) it can be trained by either forward or backward propagation; c) it is quite fast and effective, which can be trained by CPU; d) it can progressively learn feature representation.
    \item \textbf{Drawbacks:} Because MLOML is based on MOML, the performance of MLOML depends on the performance of MOML. 
    Currently, MLOML cannot efficiently handle high dimensional data well due to a full matrix $\bm{M}$ learned in MOML. This problem can be tackled by learning a diagonal matrix or employing dimensionality reduction through online feature selection, which will be investigated in the next work.
    \textcolor[rgb]{0.00,0.00,1.00}{
    Meanwhile, the number of metric layers in MLOML is uniformly specified according to the experimental results. However, the optimal number of layers is often different for different tasks. The another question is how many metric layers is sufficient for a task will be studied in the future.}
\end{itemize}

\ifCLASSOPTIONcompsoc
  \section*{Acknowledgments}
\else
  \section*{Acknowledgment}
\fi

This work is supported by the Science and Technology Innovation 2030 New Generation Artificial Intelligence Major Project (No. 2018AAA0100905), the National Natural Science Foundation of China (No. 62192783, 62106100), and the Primary Research \& Developement Plan of Jiangsu Province (No. BE2021028).

\ifCLASSOPTIONcaptionsoff
  \newpage
\fi

\appendix
\subsection{Proof of Theorem~1}


\begin{proof}
As $\bm{A}_{t+1}\!=\!(\bm{x}_{t+1}\!-\!\bm{x}_p)(\bm{x}_{t+1}\!-\!\bm{x}_p)^\top\!-\!(\bm{x}_{t+1}\!-\!\bm{x}_q)(\bm{x}_{t+1}\!-\!\bm{x}_q)^\top$, whose rank is $1$ or $2$, it has at most $2$ non-zero eigenvalues. That is to say, $\Tr(\bm{A}_{t+1})=\lambda_1+\lambda_2.$ Specifically, we can also easily get that,
\begin{equation} 
-\|\bm{x}_{t+1}-\bm{x}_q\|_2^2\leq\lambda(\bm{A}_{t+1})\leq\|\bm{x}_{t+1}-\bm{x}_p\|_2^2\,,
\end{equation}
where $\lambda(\bm{A}_{t+1})$ means the eigenvalue of $\bm{A}_{t+1}$ (\emph{i.e.,} $\lambda_1$ or $\lambda_2$). For each sample $\bm{x}$ is $\ell_2$ normalized, the ranges of $\|\bm{x}_{t+1}-\bm{x}_p\|_2^2$ and $\|\bm{x}_{t+1}-\bm{x}_q\|_2^2$ vary from $[0,4]$.
Thus,
\begin{equation} 
\lambda_{\min}(\bm{M}_t)-4\gamma\leq\lambda(\bm{M}_t-\gamma\bm{A}_{t+1})\leq\lambda_{\max}(\bm{M}_t)+4\gamma\,.
\end{equation}
When $\gamma\leq\frac{1}{4}\lambda_{\min}(\bm{M}_t)$, it is guaranteed that the minimum eigenvalue of $\bm{M}_t-\gamma\bm{A}_{t+1}$ is greater than zero. As the initial matrix $\bm{M}_1=\bm{I}$  is positive definite (\emph{i.e.,} $\lambda_{\min}(\bm{M}_1)=1$). By properly setting a small $\gamma$, the minimum eigenvalue of $\bm{M}_t-\gamma\bm{A}_{t+1}$ is generally large than zero. Thus, the positive definiteness of $\bm{M}_{t+1}=\bm{M}_t-\gamma\bm{A}_{t+1}$ can be guaranteed. Same theoretical guarantee (\emph{i.e.,} the small pertubations of positive definite matrix) can also be found in the chapter $9.6.12$ of \cite{petersen2008matrix}.
\end{proof}

\subsection{Proof of Theorem~2}

\begin{proof}
According to the objective function of MOML, \emph{i.e.,}
\begin{equation}
\Gamma=\underset{\bm{M}\succcurlyeq0}{\arg\min}\frac{1}{2}\lVert\bm{M}-\bm{M}_{t-1}\rVert_F^2+\gamma\Big[1+\Tr(\bm{M}\bm{A}_t)\Big]_+\, ,
\end{equation}
we denote $\ell_t$ as the instantaneous loss suffered by MOML at each $t$-time step with the learnt $\bm{M}_t\in\mathbb{R}^{d\times d}$, and denote by $\ell_t^\ast$ the loss suffered by an arbitrary parameter matrix $\bm{U}\in\mathbb{R}^{d\times d}$, which can be formalized as below:
\begin{equation}\label{fun_opml_L}
  \begin{split}
   \ell_t= &\ell(\bm{M}_t;\langle\bm{x}_t, \bm{x}_p, \bm{x}_q\rangle)=\lbrack 1+\Tr{(\bm{M}_t\bm{A}_t)}\rbrack_+\\
   \ell_t^\ast= &\ell(\bm{U};\langle\bm{x}_t, \bm{x}_p, \bm{x}_q\rangle)=\lbrack 1+\Tr{(\bm{U}\bm{A}_t)}\rbrack_+\,,
  \end{split}
\end{equation}
where $\bm{A}_t=(\bm{x}_t-\bm{x}_p)(\bm{x}_t-\bm{x}_p)^\top-(\bm{x}_t-\bm{x}_q)(\bm{x}_t-\bm{x}_q)^\top$, $\Tr$ denotes trace and $[z]_+=\max(0,z)$. As $\Tr(\bm{M}_t\bm{A}_t)$ is a linear function, it is convex \emph{w.r.t} $\bm{M}_t$ by natural. Besides, the hinge loss function $[z]_+$ is a convex function (but not continuous at $z=0$) \emph{w.r.t} $z$. Hence, the resulting composite function $\ell_t(\bm{M}_t)$ is convex \emph{w.r.t} $\bm{M}_t$. As $\ell$ is a convex function, we can introduce the first-order condition as follow:
\begin{equation}\label{fun_convex}
  \ell(\bm{Y})\geq\ell(\bm{X})+\VEC(\bigtriangledown\ell(\bm{X}))^\top\VEC(\bm{Y}-\bm{X})\,,
\end{equation}
where $\bm{X}, \bm{Y}\in\mathbb{R}^{d\times d}$, $\VEC$ denotes vectorization of a matrix, and $\bigtriangledown\ell(\bm{X})$ is the gradient of function $\ell$ at $\bm{X}$.

Inspired by~\cite{crammer2006online}, we define $\Delta_t$ to be $\lVert\bm{M}_t-\bm{U}\rVert_F^2-\lVert\bm{M}_{t+1}-\bm{U}\rVert_F^2$. Then calculating the cumulative sum of $\Delta_t$ over all $t\in\{1, 2, \cdots, T\}$, we can easily obtain $\sum_t\Delta_t$,
\begin{equation}
  \begin{split}
   \sum_{t=1}^T\Delta_t=&\sum_{t=1}^T(\lVert\bm{M}_t-\bm{U}\rVert_F^2-\lVert\bm{M}_{t+1}-\bm{U}\rVert_F^2)\\
                       =&\lVert\bm{M}_1-\bm{U}\rVert_F^2-\lVert\bm{M}_{T+1}-\bm{U}\rVert_F^2\\
                       \leq&\lVert\bm{M}_1-\bm{U}\rVert_F^2\,.
  \end{split}
\end{equation}

For simplicity, we employ stochastic gradient descent (SGD) to update the parameter matrix $\bm{M}_t$. Hence, according to the definition of SGD, $\bm{M}_{t+1}=\bm{M}_t-\eta\bigtriangledown\ell(\bm{M}_t)$, where $\eta$ is the learning rate, and $\bigtriangledown\ell(\bm{M}_t)=\gamma\bm{A}_{t+1}$. Then, we can rewrite the $\Delta_t$ as,
\begin{equation}
  \begin{split}
   \Delta_t=&\lVert\bm{M}_t\!-\!\bm{U}\rVert_F^2\!-\!\lVert\bm{M}_{t+1}\!-\!\bm{U}\rVert_F^2\\
           =&\lVert\bm{M}_t\!-\!\bm{U}\rVert_F^2\!-\!\lVert\bm{M}_{t}\!-\!\eta\bigtriangledown\ell(\bm{M}_t)\!-\!\bm{U}\rVert_F^2\\
           =&\lVert\bm{M}_t\rVert_F^2\!-\!2\langle\bm{M}_t, \bm{U}\rangle_F\!+\!\lVert\bm{U}\rVert_F^2
           \!-\!\lVert\bm{M}_t\!-\!\bm{U}\rVert_F^2\\
           &+2\langle\bm{M}_t\!-\!\bm{U}, \eta\bigtriangledown\ell(\bm{M}_t)\rangle_F\!-\!\eta^2\lVert\bigtriangledown\ell(\bm{M}_t)\rVert_F^2\\
           =&2\eta\VEC(\bm{M}_t\!-\!\bm{U})^\top\VEC(\bigtriangledown\ell(\bm{M}_t))\!-\!\eta^2\lVert\bigtriangledown\ell(\bm{M}_t)\rVert_F^2\\
            &\scriptsize\text{\emph{\big(employ the Eq.~(\ref{fun_convex}) i.e., }} \scriptsize\text{\emph{$\ell(\bm{U})\!\geq\!\ell(\bm{M}_t)\!+\!\VEC(\bigtriangledown\ell(\bm{M}_t))^\top\VEC(\bm{U}\!-\!\bm{M}_t)$\big)}}\\
           \geq&2\eta(\ell_t-\ell_t^\ast)-\eta^2\lVert\bigtriangledown\ell(\bm{M}_t)\rVert_F^2\,.
  \end{split}
\end{equation}
We can easily get that,
\begin{equation}
  \begin{split}
   \sum_{t=1}^T\Big[2\eta(\ell_t-\ell_t^\ast)-\eta^2\lVert\bigtriangledown\ell(\bm{M}_t)\rVert_F^2\Big]\leq\lVert\bm{M}_1-\bm{U}\rVert_F^2\,.
  \end{split}
\end{equation}

As all samples are $\ell_2$ normalized, the $2$-norm of each sample is $1$, namely $\lVert\bm{x}_t\rVert_2\equiv1, t\in\{1, 2, \cdots, T\}$. We can easily calculate the Frobenius norm of $\bm{A}_{t+1}$.
\begin{equation}\small 
  \begin{split}
   \lVert\bm{A}_{t\!+\!1}\rVert_F\!\leq\!&\lVert(\bm{x}_{t\!+\!1}\!-\!\bm{x}_p)(\bm{x}_{t\!+\!1}\!-\!\bm{x}_p)^\top\rVert_F\!+\!\lVert(\bm{x}_{t\!+\!1}\!-\!\bm{x}_q)(\bm{x}_{t\!+\!1}\!-\!\bm{x}_q)^\top\rVert_F\\
      &\scriptsize\text{\emph{\big(employ $\lVert \bm{a}\bm{b}^\top\rVert_F^2\!=\!(\sum_{i=1}^d|\bm{a}_i|^2)(\sum_{j=1}^d|\bm{b}_j|^2)$, where $\bm{a},\bm{b}\in\mathbb{R}^d$\big)}}\\
     =&\lVert\bm{x}_{t\!+\!1}\!-\!\bm{x}_p\rVert_2\cdot\lVert\bm{x}_{t\!+\!1}^\top\!-\!\bm{x}_p^\top\rVert_2\!+\!\lVert\bm{x}_{t\!+\!1}\!-\!\bm{x}_q\rVert_2\cdot\lVert\bm{x}_{t\!+\!1}^\top\!-\!\bm{x}_q^\top\rVert_2\\
     =&\lVert\bm{x}_{t+1}-\bm{x}_p\rVert_2^2+\lVert\bm{x}_{t+1}-\bm{x}_q\rVert_2^2\\
      &\scriptsize\text{\emph{\big(for $\lVert\bm{a}-\bm{b}\rVert_2^2\leq(\lVert\bm{a}\rVert_2+\lVert\bm{b}\rVert_2)^2$\big)}}\\
                         \leq&8\,.
  \end{split}
\end{equation}
Thus,
\begin{equation} 
  \begin{split}
   \sum_{t=1}^T(\ell_t-\ell_t^\ast)\leq&\frac{1}{2\eta}\lVert\bm{M}_1-\bm{U}\rVert_F^2+\frac{\eta}{2}\sum_{t=1}^T\lVert\bigtriangledown\ell(\bm{M}_t)\lVert_F^2\\
   =&\frac{1}{2\eta}\lVert\bm{M}_1-\bm{U}\rVert_F^2+\frac{\eta}{2}\sum_{t=1}^T\lVert\gamma\bm{A}_{t+1}\lVert_F^2\\
   \leq&\frac{1}{2\eta}\lVert\bm{M}_1-\bm{U}\rVert_F^2+32T\eta\gamma^2\\
     &\scriptsize\text{\emph{($\bm{M}_1$ is initialized to an identity matrix $\bm{I}$)}}\\
    =&\frac{1}{2\eta}\lVert\bm{I}-\bm{U}\rVert_F^2+32T\eta\gamma^2\,.
  \end{split}
\end{equation}

In particular, setting $\eta=\frac{1}{\Phi\sqrt{T}}$ (where $\Phi>0$ is a constant) yields the regret bound $R(\bm{U},T)\leq\big(\frac{\Phi}{2}\lVert\bm{I}-\bm{U}\rVert_F^2+\frac{32\gamma^2}{\Phi}\big)\sqrt{T}$. In fact, in this study, as a closed-form solution is employed (\emph{i.e.,} $\eta=1$), the regret bound is $R(\bm{U},T)\leq\frac{1}{2}\lVert\bm{I}-\bm{U}\rVert_F^2+32T\gamma^2$. By setting $\gamma$ in a decreasing way with the iteration number $T$, for example, $\gamma=\frac{1}{\Phi\sqrt{T}}$, we can obtain a regret bound $R(\bm{U},T)\leq\frac{1}{2}\lVert\bm{I}-\bm{U}\rVert_F^2+\frac{32}{\Phi^2}$. Hence proved.
\end{proof}

\subsection{Theoretical analysis of Proposition~1}

\begin{proof}
For simplicity, we just consider to analyze and prove this proposition of MLOML-FP that only uses forward propagation strategy. In fact, as MLOML-FP only has forward propagation, each metric layer is a relatively independent MOML algorithm. Thus, Theorem~$2$ is applicable to each metric layer. In other words, each metric layer (\emph{i.e.,} a MOML algorithm) has its own tight regret bound. As the subsequent metric layer is learnt based on the output of the former metric layer, the metric space should not be worse according to the theoretical guarantee of regret bound. Moreover, ReLU, Sigmoid, tanh activation functions can introduce nonlinear and sparsity into the feature mapping, which is also beneficial to the exploration of feature space. In some cases, if the latter metric layer is in the wrong direction, backward propagation can be chosen to correct and adjust the direction to some extent.
\end{proof}

\bibliographystyle{IEEEtran}
\bibliography{sample-bibliography}

\end{document}